\documentclass[letterpaper, 10 pt, conference]{format/ieeeconf}

\usepackage{amsmath}
\makeatletter
\def\maketag@@@#1{\hbox{\m@th\normalfont\normalsize#1}}
\makeatother
\usepackage{amsfonts}

\usepackage{amsthm}
\usepackage[ruled,vlined]{algorithm2e}
\usepackage{mathtools}
\usepackage{tabularx}
\usepackage{graphicx}
\usepackage{subfigure}
\usepackage{enumerate}
\usepackage{float}
\usepackage{booktabs}
\usepackage{url}
\usepackage{verbatim}
\usepackage[linkcolor=black,citecolor=black,urlcolor=black,colorlinks=true]{hyperref}
\usepackage{cite}
\usepackage{hyperref}
\usepackage{algorithmicx}
\usepackage{array}
\usepackage{multirow}
\usepackage{longtable}
\usepackage{rotating}
\usepackage{caption}
\usepackage{graphicx}
\usepackage{subfigure}
\usepackage{amsmath,amssymb}
\usepackage{bbm}
\usepackage{placeins}

\DeclareCaptionLabelFormat{fig}{Fig. #2}
\graphicspath{{figures/}}
\IEEEoverridecommandlockouts
\title{\LARGE \bf
Learning Human-Like Badminton Skills for Humanoid Robots	
}

\author{Yeke Chen$^{*1}$, Shihao Dong$^{*1}$, Xiaoyu Ji$^{*1}$, Jingkai Sun$^{*1}$, Zeren Luo$^{1}$, Liu Zhao$^{1}$, \\
Jiahui Zhang$^{1}$, Wanyue Li$^{1}$, Ji Ma$^{1}$, Bowen Xu$^{1}$, Yimin Han$^{1}$, \\
Xuanyi Li$^{2}$, Yudong Zhao$^{2}$, Liyun Li$^{2}$, Peng Lu$^{\dagger1}$
}

\begin{document}

\twocolumn[{%
	\renewcommand\twocolumn[1][]{#1}%
	\maketitle
	\thispagestyle{empty}
	\pagestyle{empty}

	\begin{center}
		\centering
        \vspace{-0.3cm}
		\includegraphics[width=0.98\textwidth]{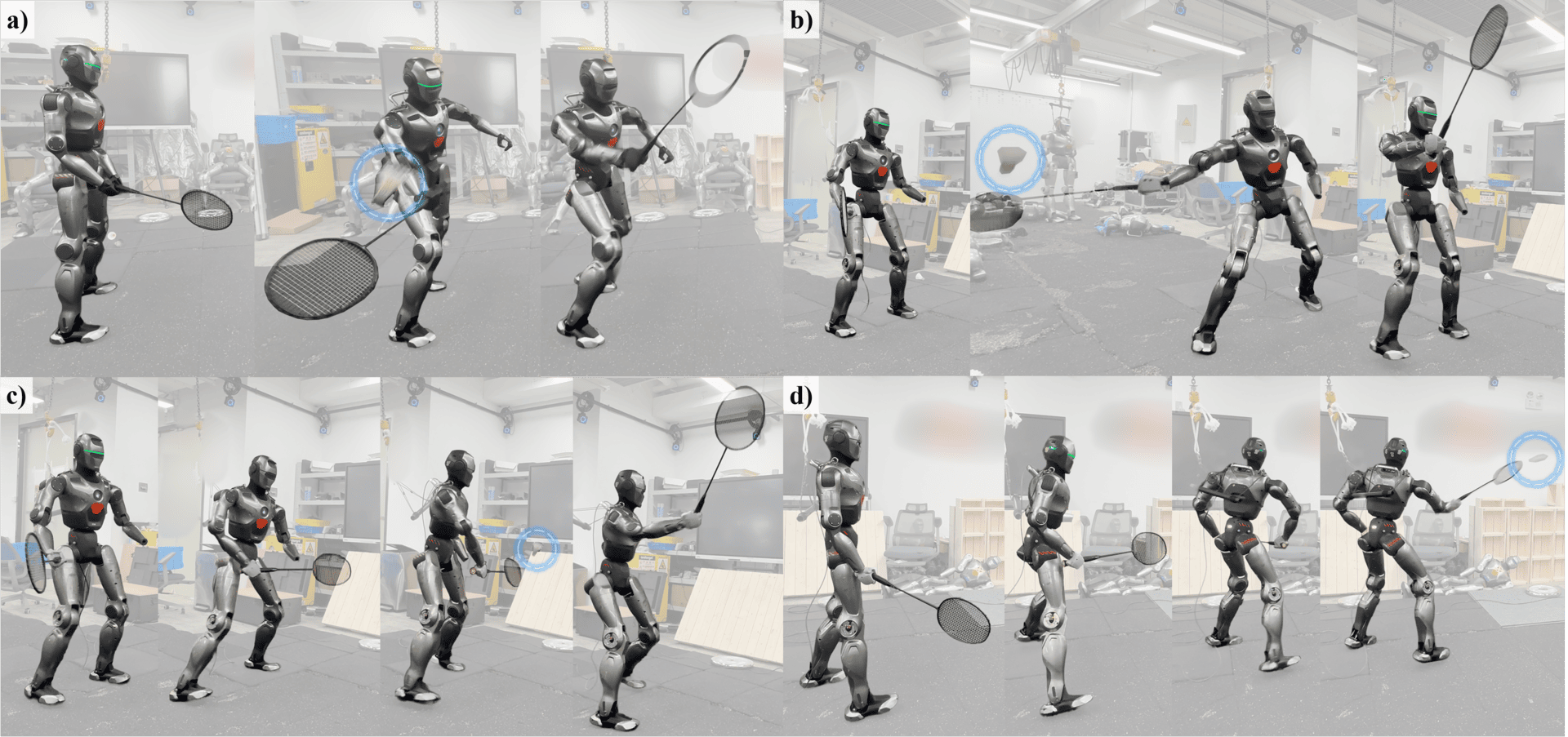}
		
		\captionsetup{skip=5pt}
		\captionsetup{font=footnotesize}
		\captionof{figure}{\textbf{Real-world Deployment of the System. }We present a learning-based framework that enables a humanoid to perform agile shuttlecock interceptions using a racket. The snapshots demonstrate the zero-shot Sim-to-Real transfer of two fundamental skills: \textbf{(a, b) Forehand Lifts} and \textbf{(c, d) Backhand Lifts}. The blue circles highlight the successful contact moments between the racket and the shuttlecock. Despite the complexity of the motions, our policy maintains robust balance and tracking accuracy on physical hardware.
		}
		\label{fig:demo}
	\end{center}%
}]

{
  \renewcommand{\thefootnote}{\relax}
  \setlength{\skip\footins}{-0.3cm} 
  \footnotetext{
    \footnotesize
    \raggedright
    \setlength{\parindent}{0pt}
    $*$ Equal contribution. $\dagger$ Corresponding author.

    $^{1}$ Authors are with the Adaptive Robotic Controls Lab (ArcLab), Department of Mechanical Engineering, The University of Hong Kong, Hong Kong SAR, China (Email: \texttt{chen-yeke@connect.hku.hk}, \texttt{lupeng@hku.hk}).
    
    $^{2}$ X. Li, Y. Zhao, and L. Li are with EngineAI Robotics Technology Co., Ltd., Shenzhen, China.
  }
}

\begin{abstract}

Realizing versatile and human-like performance in high-demand sports like badminton remains a formidable challenge for humanoid robotics. 
Unlike standard locomotion or static manipulation, this task demands a seamless integration of explosive whole-body coordination and precise, timing-critical interception. 
While recent advances have achieved lifelike motion mimicry, bridging the gap between kinematic imitation and functional, physics-aware striking without compromising stylistic naturalness is non-trivial. 
To address this, we propose \textit{Imitation-to-Interaction}, a progressive reinforcement learning framework designed to evolve a robot from a ``mimic'' to a capable ``striker.'' Our approach establishes a robust motor prior from human data, distills it into a compact, model-based state representation, and stabilizes dynamics via adversarial priors. Crucially, to overcome the sparsity of expert demonstrations, we introduce a manifold expansion strategy that generalizes discrete strike points into a dense interaction volume. 
We validate our framework through the mastery of diverse skills, including lifts and drop shots, in simulation. 
Furthermore, we demonstrate the first zero-shot sim-to-real transfer of anthropomorphic badminton skills to a humanoid robot, successfully replicating the kinetic elegance and functional precision of human athletes in the physical world.

\end{abstract}

\section{Introduction}

Badminton stands as one of the most demanding racket sports, requiring a unique combination of speed, precision, and motor diversity. A proficient human player commands a rich repertoire of strokes---from lightning-fast overhead smashes that overwhelm opponents, to delicate drop shots that barely clear the net. Each stroke involves sophisticated whole-body coordination: power originates from the legs, flows through the rotating trunk, and is ultimately delivered via a precisely timed wrist snap, forming a kinetic chain that exemplifies the elegance of human motor control.

Beyond raw performance, human badminton motions exhibit a distinctive naturalness and fluency---smooth trajectories, rhythmic timing, and biomechanically efficient postures that are immediately recognizable yet challenging to replicate. Moreover, players continuously adapt to the shuttlecock's variable trajectory, making split-second decisions on stroke selection and execution. This seamless integration of perception, decision-making, and multi-skill motor execution represents a pinnacle of human sensorimotor intelligence---and sets an ambitious target for robotic systems aspiring to achieve human-like athletic performance.

While human players seamlessly switch between thundering smashes and feathery drop shots with whole-body coordination, significant breakthroughs have been made in legged robot research, but challenges remain. Among these, \cite{liao2025beyondmimic}\cite{xie2025kungfubot}\cite{he2025asap}\cite{weng2025hdmi}\cite{yin2025visualmimic}\cite{allshire2025visual}\cite{yang2025omniretarget}\cite{wang2025physhsi} have achieved humanoid-like motions, yet lack precise interaction with tiny objects. \cite{he2025viral}\cite{ben2025homie} demonstrates successful interaction but falls short of agile, whole-body coordination. Several works have explored ball-related sports; however, some only validate their methods in simulation without real robotic system testing \cite{luo2024smplolympics}\cite{xu2025learning}\cite{yu2025skillmimic}\cite{wang2025skillmimic}\cite{zhang2023learning}, others fail to fully realize humanoid characteristics \cite{ma2025learning}\cite{liu2025humanoid}\cite{hu2025towards}\cite{nguyen2025whole}. Some have learned agile human-like motions \cite{su2025hitter}\cite{ren2025humanoid} --- yet realizing versatile and human-like performance in high-demand sports like badminton, which require precise timing, accuracy, and agility, remains a formidable challenge. These limitations highlight the gap that this work aims to address.

To bridge the gap between kinematic mimicry and dynamic interaction, we propose a progressive ``Imitation-to-Interaction'' framework. Our method decouples whole-body coordination from precise striking through a four-stage pipeline: (1) learning a robust motor prior via motion tracking; (2) performing goal-conditioned distillation to initialize a student policy with a model-based state representation (incorporating \textit{Time-to-Hit}, \textit{Target Hit State} and \textit{Target Recovery State}) and forward-compatible critics; (3) stabilizing motion execution through reinforcement learning (RL) with Adversarial Motion Priors (AMP) \cite{peng2021amp}\cite{luo2025learning}; and (4) conducting interaction-driven refinement in a physics-interactive environment. This final stage generalizes the policy from sparse dataset samples to a dense interaction manifold, transforming the agent from a ``mimic'' to a functional ``striker.'' We validate our approach through extensive experiments and successfully demonstrate agile, human-like badminton striking on a physical humanoid robot.

In summary, our main contributions are as follows:
\begin{itemize}
    \item We propose a progressive learning framework that bridges the gap between kinematic imitation and dynamic interaction. This enables the synthesis of agile, high-velocity sports movements that are distinctively \textbf{human-like} in coordination and style.
    \item We introduce a specialized state representation that \textbf{preserves} the motion priors inherent in human data. This design ensures stable transitions from kinematic tracking to physics-aware striking, maintaining natural behavioral fidelity even during aggressive maneuvers. 
    \item We achieve the first zero-shot sim-to-real transfer of whole-body badminton skills to a humanoid robot with \textbf{anthropomorphic} agility, demonstrating robust striking capabilities and stylistic realism in the physical world.
\end{itemize}

\section{Related Works}

\subsection{Humanoid Whole-Body Control} 

Achieving stable yet agile whole-body control is a fundamental challenge in humanoid robotics. Early approaches predominantly relied on model-based trajectory optimization \cite{dallard2023synchronized}\cite{dariush2008whole}\cite{ramos2019dynamic}\cite{chignoli2021humanoid}. While theoretically sound, these methods rely heavily on accurate dynamics modeling and often result in stiff behaviors that struggle to adapt to the unpredictable contact dynamics of real-world environments.

The rapid development of RL has enabled robots to learn robust controllers directly through trial-and-error\cite{rudin2022learning}\cite{zhuang2023robot}\cite{chen2025learning}\cite{luo2025mild}. To synthesize naturalistic behaviors, imitation techniques \cite{peng2018deepmimic} have been widely adopted. Seminal works like \cite{he2025asap}\cite{liao2025beyondmimic}\cite{xie2025kungfubot}\cite{zhang2025track} utilize tracking-based rewards to replicate complex kinematic skills with high fidelity. Parallel to tracking, AMP \cite{peng2021amp} provides an alternative paradigm, enabling robots to learn versatile interaction tasks while implicitly maintaining a human-like style \cite{wang2025physhsi}.

However, a disconnect remains between mimicking motion and functional interaction. While methods like \cite{yang2025omniretarget}\cite{allshire2025visual}\cite{yin2025visualmimic}\cite{weng2025hdmi}\cite{wang2025physhsi} have explored object interaction, they are often limited to large, static targets. Other works \cite{ben2025homie}\cite{he2025viral} have achieved precise manipulation of small objects via teleoperation and imitation, yet these typically focus on upper-limb dexterity, lacking the explosive, whole-body coordination required for athletic sports. Our work builds upon these foundations but ventures into a more challenging domain: leveraging human motion priors to interact with tiny, high-velocity objects in sparse-reward settings, where the demand for both precision and agility is significantly amplified.

\subsection{Dynamic Ball Sports in Legged Robotics} 

Ball sports serve as a comprehensive benchmark for testing integrated locomotion, precision control, and rapid decision-making. Significant progress has been made in ground-based tasks, such as robot soccer \cite{yin2025visualmimic}\cite{huang2023creating}, where legged agents learn robust object interaction. However, these tasks are predominantly confined to the 2D ground plane, which simplifies the interception problem compared to aerial sports.

In the realm of aerial sports, systems have been developed for badminton and table tennis using fixed robotic arms as well as quadrupedal and humanoid platforms \cite{ma2025learning}\cite{liu2025humanoid}\cite{hu2025towards}\cite{buchler2022learning}. While these works demonstrate impressive task success rates, they often prioritize functional engineering---adopting distinct, non-human strategies---over biomechanical realism. 
Concurrently, other studies have explored agile, human-like aerial ball sports \cite{xu2025learning}\cite{yu2025skillmimic}\cite{wang2025skillmimic}\cite{zhang2023learning}\cite{luo2024smplolympics}\cite{wang2024strategy}; however, these represent simulated avatars rather than physical machines, lacking validation on robotic morphologies or transfer to real-world hardware. 
Recently, several works have successfully learned agile human-like motions \cite{su2025hitter}\cite{ren2025humanoid}, enabling robots to perform dynamic maneuvers ranging from rapid table tennis strokes to agile goalkeeping actions. 
Nevertheless, realizing versatile and human-like performance in high-demand sports like badminton, which require precise timing, accuracy, and agility, remains a formidable challenge. This work aims to bridge this gap, synthesizing a policy that achieves high-performance striking without sacrificing the natural elegance and efficiency of the human kinetic chain. 

\begin{figure*} \centering 
	\hspace{-0.3cm}  
	\includegraphics[width=2.05\columnwidth]{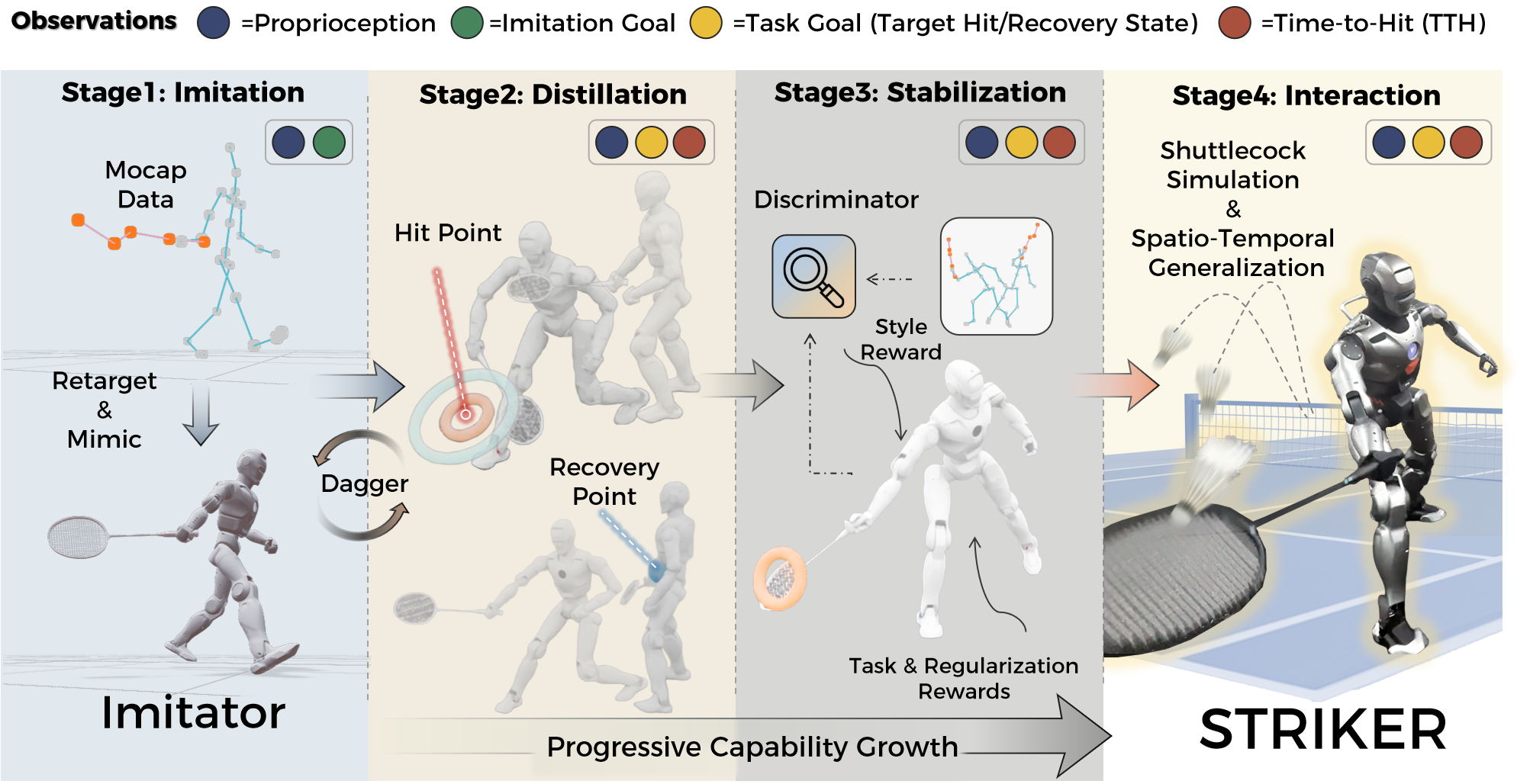} 
	\captionsetup{skip=5pt} 
	\captionsetup{font=footnotesize}  
	\caption{\textbf{Overview of the Framework.} The pipeline progressively transforms a kinematic imitator into a dynamic striker through four stages: 
\textbf{(Stage 1) Imitation:} A teacher policy learns to robustly track human motions from MoCap data using proprioceptive (blue) and imitation goal (green) observations. 
\textbf{(Stage 2) Distillation:} The teacher's capabilities are distilled into a student policy via DAgger. The student operates on a reduced observation space consisting of proprioception, task goals (yellow: target hit/recovery states), and time-to-hit (red), removing dependency on future motion trajectories.
\textbf{(Stage 3) Stabilization:} The student policy is fine-tuned using RL with an AMP discriminator to enforce stylistic plausibility (Style Reward) while minimizing tracking errors, stabilizing the motion against drift.
\textbf{(Stage 4) Interaction:} In the final physics-interactive environment, the policy undergoes refinement with simulated shuttlecock dynamics, generalizing to a dense spatio-temporal manifold to achieve precise, agile striking.
}
	\vspace{-0.5cm}  
	\label{fig:pipe}    
\end{figure*}

\subsection{Goal-Conditioned Reinforcement Learning}

Learning agile motor skills to reach dynamic targets can be formulated as a Goal-Conditioned Reinforcement Learning (GCRL) \cite{liu2022goal} problem. A fundamental challenge in this domain, particularly for high-dimensional legged robot control, is the sparsity of reward signals, which significantly complicates the exploration process \cite{li2023robust}\cite{rudin2022advanced}. 

To mitigate the exploration burden, hierarchical approaches \cite{caluwaerts2023barkour}\cite{miki2024learning}\cite{hoeller2024anymal} decouple the problem by using a high-level planner to direct a low-level locomotion controller. However, this decoupling often imposes a performance ceiling: the system's overall agility is bottlenecked by the constraints of the frozen low-level controller, limiting the emergence of coherent whole-body behaviors.

Alternatively, techniques such as Reference State Initialization (RSI) \cite{peng2018deepmimic}\cite{wu2025whole} leverage expert data to accelerate exploration by initializing the agent in valid states along reference trajectories. While RSI effectively provides high-quality starting points, it does not actively govern the transition dynamics \textit{post-initialization}. The agent still requires explicit guidance—such as tracking rewards or adversarial priors—to maintain physically plausible behavior throughout the episode.

Our work presents a novel perspective on GCRL by employing a progressive multi-stage framework. By first distilling a robust motor prior to anchoring the policy near the expert manifold and subsequently enabling interaction-driven refinement, we effectively structure the exploration process. This approach allows the agent to retain the stylistic quality of expert movements while autonomously exploiting the physics simulation to maximize task performance.

\section{Method}

\subsection{Overview}
Our framework is designed to enable a humanoid robot to master high-dynamic badminton skills by progressively bridging the gap between kinematic imitation and physics-based interaction. As illustrated in Fig. \ref{fig:pipe}, the pipeline consists of four stages: (1) \textbf{Motion Retargeting and Teacher Training}, where a privileged teacher learns to robustly track human motions; (2) \textbf{Goal-Conditioned Distillation}, where a student policy learns to execute these motions based on a hybrid state representation comprising \textit{Time-to-Hit}, \textit{Target Hit State} and \textit{Target Recovery State}; (3) \textbf{Motion Stabilization}, utilizing RL with AMP to refine the student's tracking stability; and (4) \textbf{Interaction-Driven Refinement}, where the policy is fine-tuned in a physics-interactive environment to master ball interception and recovery.

\subsection{Motion Data Processing \& Retargeting}
We collected a dataset of expert badminton motions using a motion capture system, recording the 6-DoF pose of the racket and body markers. To bridge the morphological gap between the human subject and the robot, we employ a parallel optimization-based retargeting scheme similar to \cite{kim2025pyroki}. 
The retargeting objective minimizes a weighted sum of costs:
\begin{align}
J =\ & J_{\text{global}} + J_{\text{local}} + J_{\text{ee, rotation}} + J_{\text{collision}} + J_{\text{limit}} + J_{\text{smooth}}
\end{align}
where $J_{\text{global}}$ aligns key point positions, $J_{\text{local}}$ preserves relative positions and angles between key points, $J_{\text{ee, rotation}}$ ensures the alignment of racket and feet orientation, $J_{\text{collision}}$ prevents collisions to avoid interpenetration and unnatural poses, $J_{\text{limit}}$ enforces joint limits to maintain physically plausible postures, and $J_{\text{smooth}}$ encourages motion smoothness for natural transitions. 
Crucially, we jointly optimize the root pose and joint angles to compensate for the robot's lack of spinal flexibility, ensuring a natural whole-body posture. Additionally, we simultaneously optimize the overall scale and the local scale between keypoints to reduce morphological discrepancies. 
Finally, the retargeted motions are vertically translated to align the feet with the ground plane, and contact sequences are extracted based on foot height thresholds. The hitting moments and the opponent's hitting moments (which correspond to the agent's recovery moments) are manually annotated.

\subsection{Stage 1: Kinematic Motor Prior Learning (Teacher)}

The teacher observation at timestep $t$ includes three components: proprioceptive observation, privileged observation, and tracking observation. Formally, it is defined as 
\begin{subequations}
\begin{align}
    \mathbf{o}_t^{\text{teacher}} &= \left[\mathbf{o}_t^{\text{prop}}, \mathbf{o}_t^{\text{priv}}, \mathbf{o}_t^{\text{track}}\right], \\
    \mathbf{o}_t^{\text{prop}} &= \left[\mathbf{q}_t, \dot{\mathbf{q}}_t, \boldsymbol{\omega}_t^{\text{base}}, \mathbf{g}_t^{\text{proj}}, \mathbf{a}_{t-1}\right], \\
    \mathbf{o}_t^{\text{priv}} &= \left[h_t^{\text{base}}, \mathbf{v}_t^{\text{base}}, \mathbf{c}_t^{\text{feet}}\right],  \\
    \mathbf{o}_t^{\text{track}} &= \left[\Delta \mathbf{s}_{t:t+H}, \Delta \mathbf{q}_{t:t+H}\right].
\end{align}
\end{subequations}

Here, $\mathbf{q}_t$ and $\dot{\mathbf{q}}_t$ represent the joint positions and velocities at timestep $t$; $\mathbf{v}_t^{\text{base}}$ and $\boldsymbol{\omega}_t^{\text{base}}$ denote the base linear and angular velocities; $\mathbf{g}_t^{\text{proj}}$ represents the gravity vector projected onto the base frame; $\mathbf{a}_{t-1}$ is the policy output at timestep $t-1$; $h_t^{\text{base}}$ is the base height; $\mathbf{c}_t^{\text{feet}}$ indicates binary foot contact states; $\Delta \mathbf{s}_{t:t+H}$ and $\Delta \mathbf{q}_{t:t+H}$ represent the differences between current and future root states (position, rotation, linear and angular velocity) and joint angles over the horizon $H$, respectively, all expressed in the base frame.

The reward function at timestep $t$ consists of tracking and regularization terms: 
\begin{equation}
    r_t = r_t^{\text{track}} + r_t^{\text{reg}} + r_t^{\text{term}},
\end{equation}
where the regularization reward $r_t^{\text{reg}}$ includes common terms such as energy consumption penalties and joint limit constraints, as described in \cite{margolis2023walk}.
The term $r_t^{\text{term}}$ corresponds to penalties or early termination conditions applied when the agent's state violates safety or performance constraints, such as base height dropping below a threshold, excessive base rotation, or large deviation from the reference motion. 
The tracking reward, following a similar format as in \cite{peng2018deepmimic}, can be separated into four components
\begin{equation}
    r_t^{\text{track}} = r_{\text{track, root}} + r_{\text{track, joint}} + r_{\text{track, ee}} + r_{\text{track, contact}},
\end{equation}
with $r_{\text{track, root}}$ encourages accurate tracking of the root state, while $r_{\text{track, joint}}$ promotes precise tracking of joint angles. The term $r_{\text{track, ee}}$ encourages alignment of the end-effector poses, including those of the hands, racket, and feet, and $r_{\text{track, contact}}$ fosters rhythmic and clear footwork by faithfully reproducing the ground contact schedule.
Note that the tracking rewards for the racket and the root are computed in the world coordinate frame, whereas other components, such as the hand and foot tracking, are defined in the base frame of the robot.

To improve robustness during distillation, the teacher policy is trained with heavy domain randomization and random external force perturbations. 
After the initial training, the teacher policy is rolled out to generate physically feasible reference trajectories. The teacher is then further finetuned on these rollouts to improve the precision of the trajectories, providing more accurate targets for the student policy to imitate during distillation. 

\subsection{Stage 2: Goal-Conditioned Distillation (Student)}

This stage transitions the control strategy from pure tracking to task-oriented execution. We distill the teacher's knowledge into a student policy $\pi_{S}$ using DAgger \cite{ross2011reduction}, while simultaneously employing the same critic architecture used in subsequent stages, paired with a standard RL loss to facilitate effective policy improvement.

\subsubsection{Goal-Conditioned State Representation}

We formulate the hitting task as a \textit{goal-conditioned} task, where the goal is represented using a model-based approach. Due to the rhythmic nature of badminton, the goal naturally divides into two phases: \textit{Preparation} (before impact) and \textit{Recovery} (after impact). The student observation $\mathbf{o}_{\text{student}}$ includes proprioceptive information $\mathbf{o}_t^{\text{prop}}$ and goal components:

\begin{itemize}
    \item \textbf{Time-to-Hit (TTH)}: A scalar indicating the remaining time until impact. When $TTH > 0$, the system is in the Preparation phase; when $TTH < 0$, it is in the Recovery phase.
    \item \textbf{Target Hit/Recovery State}: The relative difference between the current state and the target state (racket or root pose) at the moment of impact ($TTH=0$) or at recovery.
\end{itemize}

To mitigate out-of-distribution (OOD) issues arising from the large combinatorial state space, we implement several parallel design strategies. First, we simplify the input representation by omitting the explicit time-to-recovery variable, as it is rarely observable in practice. Second, to bound the temporal horizon, the TTH variable is explicitly clipped to the range $[-2, 2]$ seconds. Finally, a \textbf{phase-dependent masking} strategy is employed to further reduce ambiguity: when $\text{TTH} > 0$ (approach phase), recovery targets are masked to zero; conversely, when $\text{TTH} < 0$ (recovery phase), hit targets are masked.

\subsubsection{Reward Design}

While the other rewards remain consistent with the previous stage, the tracking reward is tailored for the new task, with distinct objectives in each phase. During the hitting phase ($TTH > 0$), the reward encourages accurate tracking of the racket trajectory rather than full-body imitation, weighted by an exponential decay on the absolute time-to-hit to emphasize states near impact. In the recovery phase ($TTH < 0$), the focus shifts to tracking the root pose without temporal decay, as the policy cannot reliably infer recovery timing from $TTH$. 
Formally, the rewards are defined as
\begin{subequations}
\label{eq:task_reward}
\begin{align}
r_t^{\mathrm{track, hit}} &= \exp\left(-\frac{|TTH|}{\sigma_{\mathrm{time}}}\right) \sum_{i} w_i^{\mathrm{hit}} \exp\left(-\frac{\|\Delta_i^{\mathrm{hit}}\|^2}{\sigma_i^{\mathrm{hit}}}\right), \\
r_t^{\mathrm{track, rec}} &= \mathbbm{1}(TTH < 0) \sum_{j} w_j^{\mathrm{rec}} \exp\left(-\frac{\|\Delta_j^{\mathrm{rec}}\|^2}{\sigma_j^{\mathrm{rec}}}\right),
\end{align}
\end{subequations}
where $i$ and $j$ index state components for hitting and recovery phases, respectively. The manifold-aware error is
\begin{equation}
\Delta_\ell^{(\cdot)} := s_\ell^{\mathrm{ref}} \boxminus s_\ell,
\end{equation}
representing the difference between the reference $s_\ell^{\mathrm{ref}}$ and current state $s_\ell$. The squared norm sums over the state dimension:
\begin{equation}
\|\Delta_\ell^{(\cdot)}\|^2 = \sum_m (\Delta_{\ell,m}^{(\cdot)})^2.
\end{equation}

\subsection{Stage 3: Motion Stabilization with RL}
While the distillation stage initializes the general policy behavior, the supervised DAgger loss inherently prioritizes global pose imitation, forcing a trade-off that often yields suboptimal racket tracking precision. To bridge this gap, this stage utilizes pure trial-and-error optimization to shift the focus towards high-precision racket tracking. 
This enables the robot to enhance its physical stability and minimize interception errors. 

Essentially, in this stage, the student policy is fine-tuned with Adversarial Motion Priors (AMP) \cite{peng2021amp}. To enforce stylistic plausibility, we introduce a discriminator network $D$ trained to distinguish between the agent's generated transitions and ground-truth motions. The discriminator operates on a specific observation space $\mathbf{o}^{\text{amp}}_t$, which aggregates a history of $H_{AMP}=5$ frames to capture temporal motion characteristics:
\begin{equation}
    \mathbf{o}^{\text{amp}}_t = [\mathbf{x}_t, \mathbf{x}_{t-1}, \dots, \mathbf{x}_{t-H_{AMP}+1}]
\end{equation}
The single-frame feature vector $\mathbf{x}_t$ consists of kinematic states expressed in the robot's local base frame:
\begin{equation}
    \mathbf{x}_t = \left[ \mathbf{v}_t^{\text{base}}, \mathbf{q}_t, h_t^{\text{base}}, \mathbf{g}_t^{\text{proj}}, \mathbf{p}_t^{\text{ee}}, \mathbf{v}_t^{\text{ee}} \right]
\end{equation}
Here, $\mathbf{p}_t^{\text{ee}}$ and $\mathbf{v}_t^{\text{ee}}$ correspond to the positions and linear velocities of the key end-effectors (ankles and hands) in the base frame. 

The optimization objective follows the Least-Squares GAN (LSGAN) formulation with a gradient penalty \cite{escontrela2022adversarial}. The total reward $r_t$ is a weighted combination of the task reward and the style reward:
\begin{equation}
    r_t = w_g r^{\text{task}}_t + w_s \max \left(0, 1 - 0.25 (D(\mathbf{o}^{\text{amp}}_t) - 1)^2 \right)
\end{equation}
where $r^{\text{task}}_t$ adopts the formulation in Eq~(\ref{eq:task_reward}).

The discriminator minimizes the following loss:
\begin{equation}
\begin{aligned}
    \mathcal{L}_D = \; & \mathbb{E}_{\mathbf{o}^{\text{D}}_t \sim \mathcal{M}} [(D(\mathbf{o}^{\text{D}}_t) - 1)^2] + \mathbb{E}_{\pi} [(D(\mathbf{o}^{\text{amp}}_t) + 1)^2] \\
    & + \frac{w_{\text{gp}}}{2} \mathbb{E}_{\mathbf{o}^{\text{D}}_t \sim \mathcal{M}} [\|\nabla D(\mathbf{o}^{\text{D}}_t)\|^2]
\end{aligned}
\end{equation}
where $\mathcal{M}$ represents the reference motion dataset and the last term is the gradient penalty regularizer applied to referenced data samples $\mathbf{o}^{\text{D}}_t$.

\subsection{Stage 4: Interaction-Driven Refinement}

In the final stage, we introduce the physics of the shuttlecock. The policy interacts with a physically simulated ball, allowing it to refine its control based on realistic feedback and improve return effectiveness.

\subsubsection{Manifold Expansion}

\begin{figure} \centering 
	\hspace{-0.3cm}  
	\includegraphics[width=0.97\columnwidth]{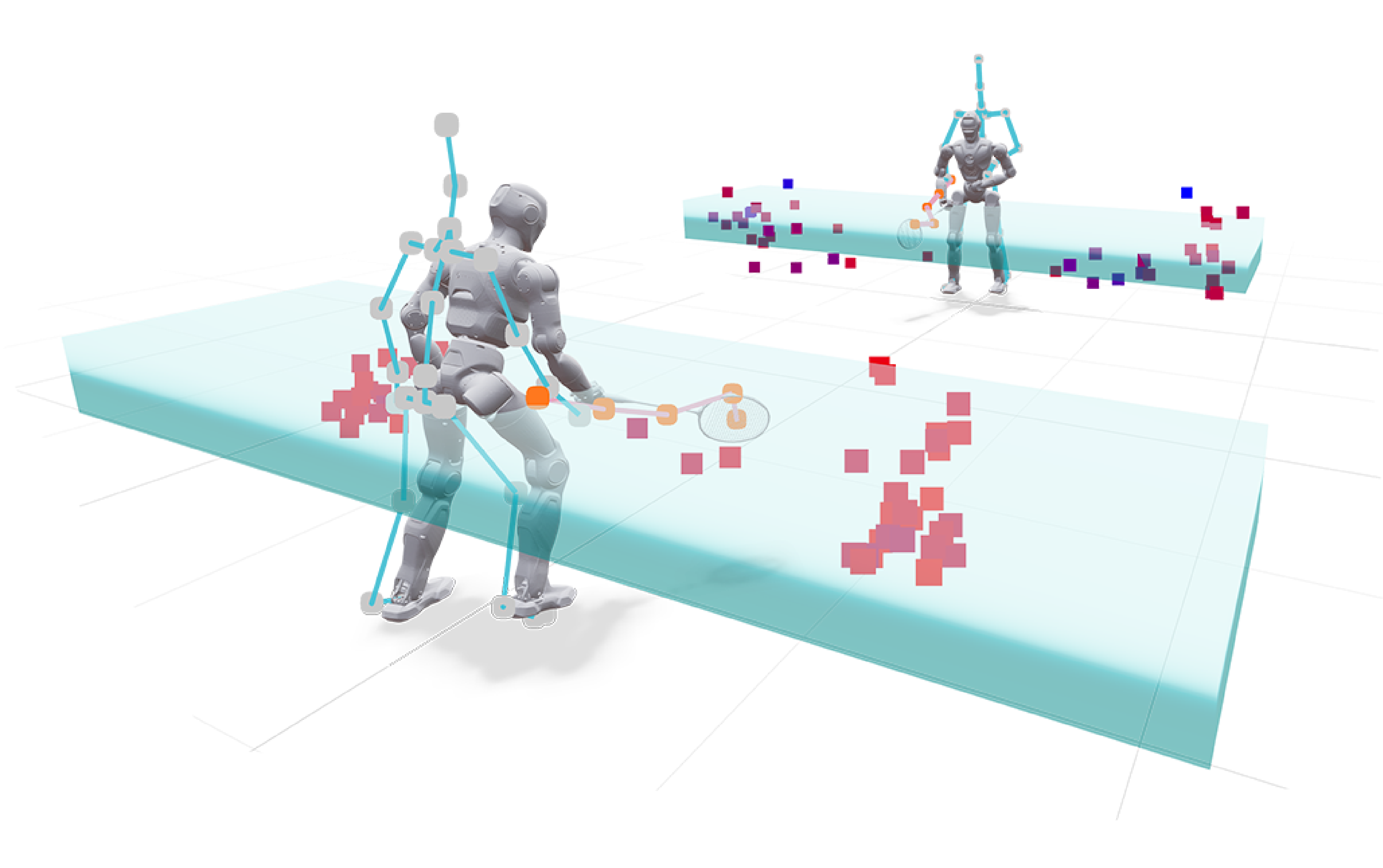} 
	\captionsetup{skip=5pt} 
	\captionsetup{font=footnotesize}  
	\caption{\textbf{Manifold Expansion of Strike Targets.} The scattered points represent the discrete strike locations from the original MoCap dataset, color-coded by the time-to-hit (blue for shorter, red for longer durations) relative to the robot's initial pose. The semi-transparent cyan volume illustrates the expanded, continuous striking manifold achieved by our interaction-driven refinement stage. Our method empowers the robot to generalize from these sparse human demonstrations to a dense, volumetric region of reachable targets across varying temporal horizons.
    }
	\vspace{-0.5cm}  
	\label{fig:hits}    
\end{figure}

Due to the sparsity of strike points in the original Mocap dataset, we expand these discrete points into a denser, near-continuous manifold of reachable strike targets, as shown in Fig. \ref{fig:hits}. Concretely, we pre-generate a large number of shuttlecock trajectories with varying interception heights and times, then select strike points around the original dataset samples. During training, the policy is tasked with generalizing motor primitives to this enriched target space.

\subsubsection{Reward Design}

We maintain the tracking rewards from the previous stage for both the hitting and recovery phases. However, the time-decay weighting in the hitting tracking reward is replaced by a sparse indicator function around the exact strike moment:
\[
r_t^{\mathrm{track, hit}} = \mathbbm{1}(|TTH| < \epsilon) \sum_i w_i^{hit} \exp\left(-\frac{\|\Delta^{hit}_i\|^2}{\sigma^{hit}_i}\right),
\]
where \(\epsilon > 0\) is a small threshold defining a narrow time window near impact. This change reflects the fact that only the strike point is known precisely, while racket trajectories immediately before and after are uncertain. Consequently, the policy learns to focus on achieving accurate motion at the strike instant. For the recovery phase, the tracking reward remains unchanged. 

Additionally, we introduce a physics-informed reward based on the ball's post-hit behavior: 
\[
r_t^{\text{hit}} = r_t^{\text{dir}} \cdot r_t^{\text{speed}},
\]
where \(r_t^{\text{dir}}\) encourages the ball to land within the opponent's court, and \(r_t^{\text{speed}}\) penalizes weak returns to promote effective strikes.

\subsubsection{Physics Simulation of the Ball}

The shuttlecock's flight is simulated by applying aerodynamic forces, including air drag and damping torques, to capture its overall flight behavior. In contrast, the contact simulation models the shuttlecock as two distinct parts—the head and the feather skirt—each assigned different material properties. This separation allows for more accurate and physically plausible collision responses with the racket and environment, enabling the policy to learn refined hitting strategies that account for both aerodynamic and contact dynamics.

\subsubsection{Temporal Rhythm Randomization}

To enhance the policy's adaptability to variable rhythmic patterns, we implement a stochastic serving and reset mechanism. Specifically, the time interval between a hitting event and the subsequent serve is randomized uniformly within the range of $[1, 6]$ seconds. This design exposes the robot to a diverse set of temporal states during training. The longer intervals compel the robot to maintain stable idling behaviors in the absence of immediate targets, ensuring energy efficiency and balance. Conversely, the shorter intervals simulate continuous rallies, requiring the robot to intercept incoming shots even when it has not yet fully recovered to a neutral stance, thereby enforcing robustness against suboptimal initial states. 

By integrating manifold expansion, physics-based ball simulation, temporal rhythm randomization, and refined rewards, Stage 4 enables the policy to transition from purely kinematic tracking to interaction-aware control. 

\begin{figure*} \centering 
	\hspace{-0.3cm}  
	\includegraphics[width=1.97\columnwidth]{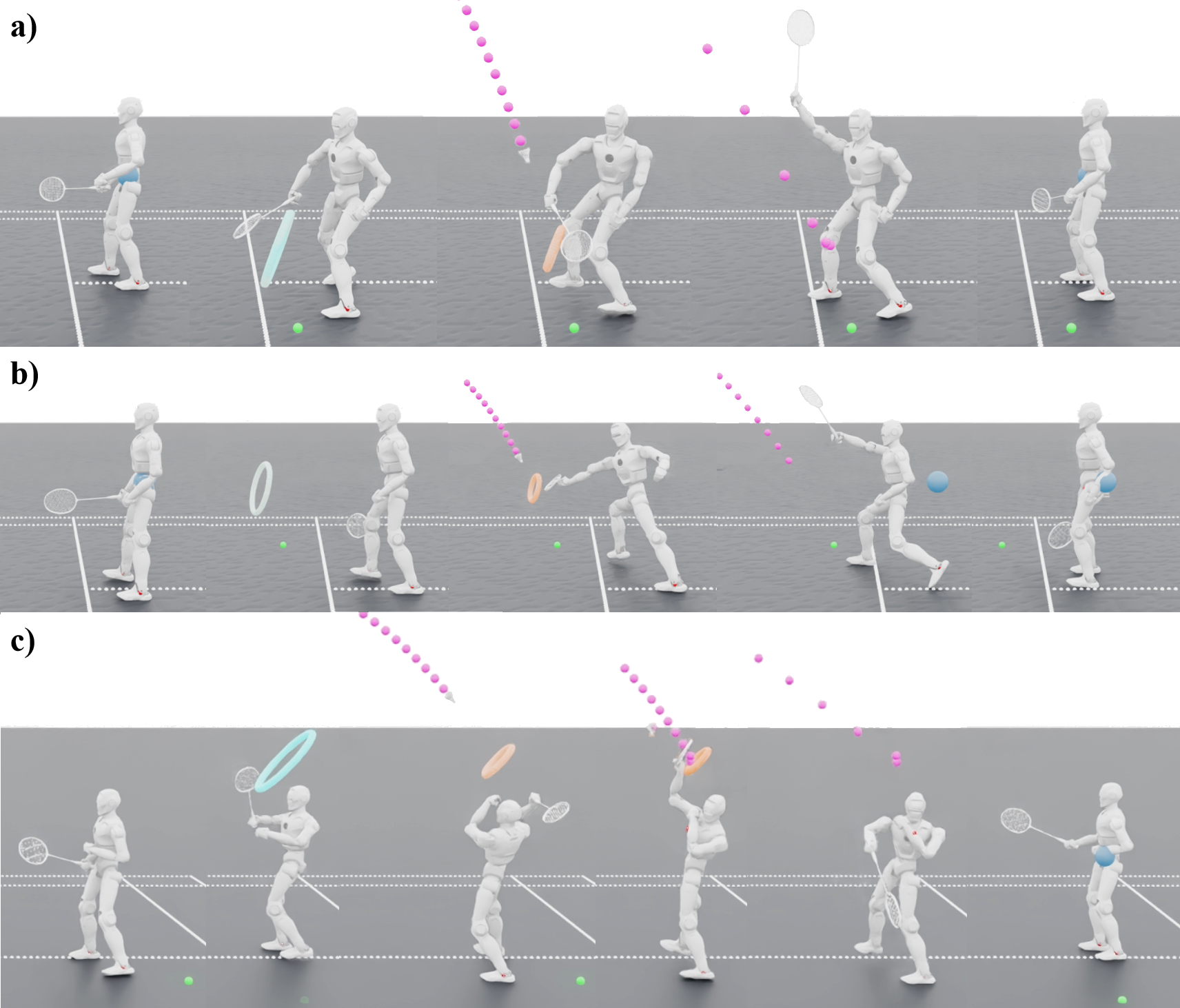} 
	\captionsetup{skip=5pt} 
	\captionsetup{font=footnotesize}  
 \caption{\textbf{Diverse Badminton Skills Learned via the Proposed Framework.} Time-lapse sequences demonstrating the humanoid's mastery of distinct striking techniques. 
\textbf{(a) Backhand Lift:} The robot rotates its torso to generate ``whipping'' power for a cross-body return.
\textbf{(b) Forehand Lift:} The robot executes an extended lunge to reach a distant target and quickly recovers balance.
\textbf{(c) Drop Shot:} The robot performs an overhead strike with a natural inertial follow-through.
Pink dots visualize the historical trajectory of the shuttlecock.}

\label{fig:skills_sim}
\end{figure*}

\section{Experimental Results}
\subsection{Implementation Details} 
\label{subsec:implementation}

\paragraph{Simulation and Training Framework.}
We develop our training environment using NVIDIA Isaac Sim, leveraging the Isaac Lab to facilitate high-throughput parallel reinforcement learning. The physics engine operates at a frequency of $200\,\text{Hz}$, while the policy runs at $50\,\text{Hz}$. Within this simulation, we successfully train the humanoid to master a diverse repertoire of badminton skills without modifying the core pipeline, specifically including forehand and backhand lifts and drop shots, as visualized in Fig.~\ref{fig:skills_sim}. 

\paragraph{Hardware and Perception Setup.}
We conduct real-world evaluations using the EngineAI PM01 humanoid robot. To obtain precise ground truth states for the experiments, we utilize an FZMotion optical motion capture system. This system tracks the 6-DoF pose of the robot's base link as well as the real-time 3D position of the badminton ball. To handle sensor noise and latency, we implement an Extended Kalman Filter (EKF) that estimates the ball's full state (position and velocity) from the raw motion capture observations, which is also employed to predict the future trajectory and the optimal hitting point, enabling the robot to plan interception motions proactively.

\paragraph{Domain Randomization.}
To facilitate robust Sim-to-Real transfer and ensure the policy remains effective under real-world uncertainties, we apply extensive domain randomization during the training phase. 
The specific randomization ranges for these critical parameters are detailed in Table~\ref{tab:rand}.

\begin{table}[h]
	\centering
	\captionsetup{skip=5pt}
	\captionsetup{font=footnotesize}
	\begin{tabular}{lcc}
	\toprule
	\textbf{Parameters} & \textbf{Range} & \textbf{Unit} \\
	\midrule
	\multicolumn{3}{l}{\textit{Robot Dynamics \& Payload}} \\
	Base Mass Variation & $[-3.0, 5.0]$ & $\text{kg}$ \\
    Hand Mass Variation & $[-0.05, 0.15]$ & $\text{kg}$ \\
    Racket Mass Variation & $[-0.005, 0.005]$ & $\text{kg}$ \\
	CoM Offset $(x, y)$ & $[-0.05, 0.05]$ & $\text{m}$ \\
	CoM Offset $(z)$ & $[-0.03, 0.03]$ & $\text{m}$ \\
	PD Gain Scale & $[0.9, 1.1]$ & - \\
    Control Latency & $[5, 30]$ & $\text{ms}$ \\
	\midrule
	\multicolumn{3}{l}{\textit{Environment \& Interaction}} \\
	Ground Friction & $[0.5, 1.0]$ & - \\
	Restitution Coeff. & $[0.0, 0.2]$ & - \\
	Base Velocity Perturbation & $[-0.4, 0.4]$ & $\text{m/s}$ \\
	Terrain Height Noise & $[0, 0.05]$ & $\text{m}$ \\
	\bottomrule
	\end{tabular}
    \caption{\textbf{Domain Randomization Parameters.} We randomize dynamic and environmental parameters to bridge the sim-to-real gap.}
    \label{tab:rand}
	\vspace{-0.3cm}
\end{table}

\begin{table*}[h]
\centering
\captionsetup{skip=5pt}
\captionsetup{font=footnotesize}
\begin{tabular}{lcccccc}
\toprule
Method & \multicolumn{2}{c}{SR$\uparrow$} & \multicolumn{2}{c}{MSE$\downarrow$} & \multicolumn{2}{c}{IBR$\uparrow$} \\
\cmidrule(lr){2-3}
\cmidrule(lr){4-5}
\cmidrule(lr){6-7}
 & easy & hard & easy & hard & easy & hard \\
\midrule
Ours & \textbf{0.9516 $\pm$ 0.0005} & \textbf{0.9153 $\pm$ 0.0007} & \textbf{0.0062 $\pm$ 0.0003} & \textbf{0.0108 $\pm$ 0.0003} & 0.1999 $\pm$ 0.0037 & \textbf{0.1575 $\pm$ 0.0042} \\
w/o Stab. & 0.9501 $\pm$ 0.0011 & 0.9147 $\pm$ 0.0016 & 0.0094 $\pm$ 0.0001 & 0.0162 $\pm$ 0.0003 & 0.1363 $\pm$ 0.0039 & 0.1148 $\pm$ 0.0031 \\
E2E-AMP & 0.8586 $\pm$ 0.0016 & 0.7391 $\pm$ 0.0016 & 0.2109 $\pm$ 0.0024 & 0.6004 $\pm$ 0.0006 & \textbf{0.2383 $\pm$ 0.0061} & 0.0990 $\pm$ 0.0002 \\
w/o Interact. & 0.3787 $\pm$ 0.0017 & 0.3583 $\pm$ 0.0014 & 0.3419 $\pm$ 0.0068 & 0.3569 $\pm$ 0.0013 & -0.0210 $\pm$ 0.0006 & -0.0230 $\pm$ 0.0008 \\
ASE-Based & 0.0037 $\pm$ 0.0006 & 0.0034 $\pm$ 0.0000 & 4.2742 $\pm$ 0.0422 & 4.3395 $\pm$ 0.0000 & -0.2671 $\pm$ 0.0003 & -0.2682 $\pm$ 0.0000 \\
VQ-Based & 0.0042 $\pm$ 0.0003 & 0.0041 $\pm$ 0.0003 & 3.3111 $\pm$ 0.0329 & 3.3766 $\pm$ 0.0172 & -0.2770 $\pm$ 0.0009 & -0.2753 $\pm$ 0.0032 \\
\bottomrule
\end{tabular}
\caption{\textbf{Quantitative Comparison in Simulation. } 
We report the performance of our method against baselines on the Lift task. Best results are highlighted in \textbf{bold}.
}
\label{tab:sim}
\end{table*}

\subsection{Simulation Experiments}
\label{subsec:sim_exp}

To comprehensively evaluate the efficacy of our proposed framework, we conduct extensive experiments in simulation. The primary goal is to verify the contribution of each stage in our pipeline and to demonstrate its superiority over representative end-to-end and hierarchical baselines.

\paragraph{Task and Environment Setup}

We focus on the Lift task, evaluated in a specialized environment designed to test the robot's interception capabilities. We classify the evaluation scenarios into two difficulty levels based on the spatial extent of the required coverage. Both modes define a total effective striking volume that encompasses both \textit{forehand} and \textit{backhand} areas relative to the robot's base:
\begin{itemize}
    \item \textbf{Easy Mode:} The shuttlecock targets are sampled within a moderate volume of $2.0 \times 0.4 \times 0.3\,\text{m}^3$.
    \item \textbf{Hard Mode:} The target volume is expanded to $4.0 \times 1.0 \times 0.3\,\text{m}^3$, necessitating agile bi-directional locomotion and versatile whole-body coordination to cover the substantial lateral range. 
\end{itemize}
Statistical results are aggregated over $4,096$ parallel environments, with each episode lasting $4,000$ steps ($80$ seconds).

\vspace{0.3cm}

\paragraph{Baselines and Ablations}

We compare our full method (\textbf{Ours}) against the following ablations and baselines:
\begin{itemize}
    \item \textbf{w/o Stab.:} Removes the motion stabilization stage (Stage 3). The policy distilled from the teacher (Stage 2) is directly fine-tuned in the interaction environment.
    \item \textbf{E2E-AMP:} Removes the imitation stages (Stages 1-3). The policy learns from scratch using AMP and task rewards directly in the physics environment.
    \item \textbf{w/o Interact.:} Removes the final interaction-driven refinement (Stage 4). We evaluate the policy immediately after the motion stabilization stage.
    \item \textbf{ASE-Based:} A hierarchical approach based on Adversarial Skill Embeddings (ASE) \cite{peng2022ase}, utilizing a latent space to drive low-level control, like \cite{wang2024strategy}. 
    \item \textbf{VQ-Based:} A hierarchical approach utilizing a VQ-VAE-based latent space \cite{zhu2023neural} for action generation.
\end{itemize}

\vspace{0.3cm}

\paragraph{Evaluation Metrics}

We quantify the performance using three key indicators:
\begin{itemize}
    \item \textbf{Success Rate (SR):} The percentage of episodes where the robot successfully intercepts and contacts the shuttlecock, measuring the fundamental reliability of the policy.
    \item \textbf{Tracking Error (MSE):} The mean squared Euclidean distance between the racket's sweet spot and the ball's position at the moment of impact. This metric evaluates the fine-grained spatial precision of the control.
    \item \textbf{In-Bounds Reward (IBR):} A continuous cumulative metric reflecting the overall quality of the return shot. Unlike a binary success count, IBR aggregates weighted rewards for valid landings within court boundaries while penalizing out-of-bounds or net collisions.
\end{itemize}

\textit{Importance of Progressive Stages:}
\textbf{Ours} achieves the best overall performance across both difficulty levels, particularly in SR and MSE. Comparing \textbf{Ours} with \textbf{E2E-AMP}, we observe a significant performance drop in the latter. This indicates that the teacher mimicry and distillation stages (Stages 1 \& 2) are critical for establishing a valid motor prior; without them, the RL agent struggles to explore all effective striking poses. 
Interestingly, \textbf{E2E-AMP} achieves a higher IBR than \textbf{Ours} in the \textit{Easy} setting. This is a case of survivorship bias: the E2E policy has a lower Success Rate and only intercepts the most accessible balls (those easy to return in-bounds), whereas \textbf{Ours} aggressively attempts to return difficult shots, occasionally leading to out-of-bounds penalties.

\textit{Role of Stabilization and Refinement:}
\textbf{w/o Stab.} achieves performance metrics comparable to the full method, indicating that the motion stabilization stage is not the primary determinant of the final performance. However, its contribution is critical for the development efficiency. In practice, a significant portion of the workflow is dedicated to tuning the interaction environment and reward functions. By providing a robust policy initialization, Stage 3 drastically reduces the convergence time required for each tuning iteration, thereby streamlining the overall experimentation and debugging process. 
Conversely, \textbf{w/o Interact.} exhibits poor performance, even yielding negative IBR in some cases. This highlights a critical gap between kinematic fidelity and functional success: merely mimicking human poses does not guarantee physically valid shuttlecock trajectories (e.g., ensuring the ball clears the net). Furthermore, the substantially low success rate without Stage 4 underscores that the sparse MoCap data is insufficient for generalization across dynamic interception scenarios.

\textit{Comparison with Hierarchical Baselines:}
Both \textbf{ASE-Based} and \textbf{VQ-Based} methods struggle significantly in this task. The ASE-based approach suffers from mode collapse; faced with the strict hardware constraints of the humanoid (limited torque, heavy limbs) and domain randomization, the low-level controller conservatively converges to a narrow set of stable poses, failing to execute the dynamic reaches required for badminton. Similarly, the VQ-Based method fails to achieve precise control, likely because the compressed discrete latent space lacks the high-frequency granularity needed for fine-grained interception tasks.

\vspace{0.3cm}

\paragraph{Qualitative Analysis of Learned Behaviors}

Beyond quantitative metrics, we observe that the learned policy exhibits sophisticated, non-conservative movement patterns that align with professional badminton biomechanics. As visualized in Fig.~\ref{fig:skills_sim}, the robot does not treat the racket as a static paddle; instead, it learns to execute complex \textbf{slicing motions} where the racket's velocity vector is intentionally non-orthogonal to the racket face. This allows for precise control over the shuttlecock's trajectory, rather than simple elastic collisions.

Specific emergent behaviors highlight the policy's mastery of whole-body dynamics:
\begin{itemize}
    \item \textbf{Whole-Body Torque Generation (Fig.~\ref{fig:skills_sim}a):} In the backhand lift, the robot initiates a distinct torso rotation away from the net. During the strike, it rapidly ``uncoils'' its trunk, transferring angular momentum from the core to the arm. This whipping motion allows the robot to generate sufficient return force to ``twist'' the shuttlecock back to the opponent's court, compensating for the limited range of motion in the robot's wrist.
    
    \item \textbf{Dynamic Lunge and Recovery (Fig.~\ref{fig:skills_sim}b):} Faced with a distant forehand target, the robot performs an aggressive lunge, shifting its Center of Mass (CoM) significantly forward to extend its reach. Crucially, immediately after the strike, the policy utilizes ground reaction forces to push back, rapidly recovering to a neutral stance. This ``attack-and-return'' pattern demonstrates the policy's foresight in managing stability for subsequent shots.
    
    \item \textbf{Inertial Follow-Through (Fig.~\ref{fig:skills_sim}c):} During the drop shot, the robot adopts a high preparatory posture (``cocking'' phase) before accelerating the racket for a high-velocity overhead strike. Notably, the arm does not stop abruptly at the impact point; instead, the policy exhibits a natural follow-through, allowing the arm to swing downward, dissipating the kinetic energy. This smooth deceleration mimics human motor control strategies to reduce joint stress and maintain fluid motion.
    
\end{itemize}

\subsection{Real-World Experiments}
\label{subsec:real_world}

We conduct real-world deployment to validate the feasibility of our zero-shot Sim-to-Real transfer. Specifically, we deploy both the learned lift policy directly onto the EngineAI PM01 humanoid robot without any fine-tuning on the physical hardware.

\paragraph{Qualitative Analysis: Whole-Body Coordination.}
As illustrated in Fig.~\ref{fig:demo}, for both forehand and backhand lift skills, the robot faithfully reproduces the temporal structure of a professional stroke. 
Specifically, in the \textbf{Backhand Lift} (Fig.~\ref{fig:demo}c-d), we observe a pronounced trunk rotation: the robot turns its back partially towards the net to clear space for the arm to cross the body, generating torque through sudden uncoiling. 
In the \textbf{Forehand Lift} (Fig.~\ref{fig:demo}a-b), the robot executes a deep lunge to extend its reach. Notably, despite the high velocity of the swing, the robot maintains stability through the follow-through phase, dissipating the arm's kinetic energy while actively adjusting its foot placement to recover a neutral stance.

A critical hardware constraint of the PM01 robot is that its arms possess only \textbf{5 Degrees of Freedom (DoF)}, lacking the full wrist dexterity used by humans. To overcome this, our policy successfully exploits whole-body coordination. By coupling the arm motion with torso yaw and precise base orientation, the robot aligns its entire body to compensate for the missing arm DoFs, ensuring the racket face strikes the shuttlecock at the correct angle to propel it forward.

\paragraph{Sim-to-Real Gap Analysis.}
While the robot successfully performs the tasks, we observe distinct behavioral shifts compared to simulation due to the inevitable reality gap. Unlike the smooth, single-motion lunges in simulation, the physical robot exhibits high-frequency footwork adjustments (i.e., rapid shuffling steps) during the approach and recovery phases. 
We attribute this primarily to the system latencies and unmodeled joint friction, which cause slight tracking delays. The policy detects these discrepancies and reacts by generating rapid corrective steps to maintain the CoM within the support polygon. 
Crucially, despite these oscillations and the lack of a perfectly stable visual appearance, the robot does not fall and successfully intercepts the ball. This highlights the policy's robustness: rather than overfitting to an ideal physics model, it adapts to dynamic uncertainties by prioritizing balance recovery over kinematic smoothness.

\paragraph{Quantitative Results.}
We evaluated the skills in a controlled test set of 10 trials each. The policy achieved a success rate of $90\%$ for forehand lifts and $70\%$ for backhand lifts. These results confirm that our framework effectively synthesizes dynamic motion primitives that are robust enough for real-world deployment despite hardware limitations.

\section{Conclusion}
\label{sec:conclusion}

In this work, we presented \textit{Imitation-to-Interaction}, a progressive reinforcement learning framework enabling humanoid robots to master dynamic badminton skills. By bridging the gap between kinematic human demonstrations and physics-based interactions, we successfully demonstrated diverse skills in simulation and validated the feasibility of zero-shot transfer on a physical humanoid robot.

However, a primary limitation remains in the delicate trade-off between \textbf{motion naturalness, dynamic stability, and task precision}. We observe that these objectives often compete directly against one another: pursuing human-like aesthetics (e.g., deep lunges and torso rotations) introduces instability, particularly under real-world latencies; prioritizing pure interception accuracy can lead to physiologically implausible (``twisted'') postures; while excessive stability constraints result in conservative behaviors that degrade agility. Currently, balancing these conflicting objectives requires meticulous tuning.
Additionally, our current system focuses on intercepting single shots within a localized area. Achieving continuous \textbf{human-robot rallies} and scaling the locomotion to support \textbf{full-court coverage} remain open challenges. 

Future work will aim to address the stability-agility conflict through more adaptive control frameworks and extend the robot's capabilities to enable sustained, full-court gameplay against human opponents.

\section{Acknowledgment}

This work was conducted during the author's internship at EngineAI. The authors would like to thank EngineAI for the support and resources provided. We also extend our gratitude to Yinghui Li, Erdong Xiao, Guanda Li and Zhilin Xiong for their insightful discussions and valuable feedback throughout this project.

\newlength{\bibitemsep}\setlength{\bibitemsep}{0.00\baselineskip}
\newlength{\bibparskip}\setlength{\bibparskip}{0pt}
\let\oldthebibliography\thebibliography
\renewcommand\thebibliography[1]{
	\oldthebibliography{#1}
	\setlength{\parskip}{\bibitemsep}
	\setlength{\itemsep}{\bibparskip}
}

\bibliography{references}

\end{document}